\documentclass[review]{elsarticle}

\usepackage{lineno,hyperref}
\modulolinenumbers[5]
\usepackage{amsmath}
\usepackage{arabtex}
\usepackage{utf8}
\usepackage{algorithm}
\usepackage{eqparbox}
\usepackage{algorithmicx}
\usepackage{algpseudocode}
\def\endabstract{\egroup}

\usepackage[utf8]{inputenc} 
\usepackage{url}            
\usepackage{booktabs}       
\usepackage{amsfonts}       
\usepackage{nicefrac}       
\usepackage{microtype}      
\usepackage{multirow}
\usepackage{multicol}
\usepackage{graphicx}
\usepackage{xargs} 
\usepackage{amsthm}
\usepackage{caption}

\usepackage{varwidth}
\DeclareCaptionFormat{myformat}{%
  \begin{varwidth}{\linewidth}%
    \centering
    #1#2#3%
  \end{varwidth}%
}

\newcommand{\algrule}[1][.2pt]{\par\vskip.5\baselineskip\hrule height #1\par\vskip.5\baselineskip}

\hypersetup{
    colorlinks,
    linkcolor=cyan,
    citecolor = cyan,
    filecolor=cyan,      
    urlcolor=cyan,
    }

\algnewcommand\algorithmicforeach{\textbf{for}}
\algdef{S}[FOR]{For}[1]{\algorithmicfor\ \#1\ \algorithmicdo}

\biboptions{authoryear}

\begin{document}
\begin{frontmatter}
\title{An End-to-End OCR Framework for Robust Arabic-Handwriting Recognition using a Novel Transformers-based Model and an Innovative 270 Million-Words Multi-Font Corpus of Classical Arabic with Diacritics}
\author[1]{Aly Mostafa \corref{mycorrespondingauthor}\fnref{eq}}\ead{alymostafa@fci.helwan.edu.eg}\cortext[mycorrespondingauthor]{Corresponding author}
\fntext[eq]{Equal Contributions}
\author[1]{Omar Mohamed\fnref{eq}}\ead{omar_20170353@fci.helwan.edu.eg}  
\author[1]{Ali Ashraf\fnref{eq}}\ead{aliashraf@fci.helwan.edu.eg}
\author[1]{Ahmed Elbehery\fnref{eq}}\ead{ahmedismail@fci.helwan.edu.eg}
\author[1]{Salma Jamal\fnref{eq}}\ead{salmagamal@fci.helwan.edu.eg}
\author[1]{Anas Salah}\ead{Anas\_20170123@fci.helwan.edu.eg}
\author[1]{Amr S. Ghoneim}\ead{amr.ghoneim@fci.helwan.edu.com}
\address[1]{Departement of Computer Science, Faculty of Computers and Artificial Intelligence, Helwan University, Helwan, Egypt}
\begin{abstract}
This research is the second phase in a series of investigations on developing an Optical Character Recognition (OCR) of Arabic historical documents and examining how different modeling procedures interact with the problem. The first research studied the effect of Transformers on our custom\-built Arabic dataset. One of the downsides of the first research was the size of the training data, a mere 15000 images from our 30 million images, due to lack of resources. Also, we add an image enhancement layer, time and space optimization, and Post\-Correction layer to aid the model in predicting the correct word for the correct context. Notably, we propose an end\-to\-end text recognition approach using Vision Transformers as an encoder, namely BEIT, and vanilla Transformer as a decoder, eliminating CNNs for feature extraction and reducing the model's complexity. The experiments show that our end\-to\-end model outperforms Convolutions Backbones. The model attained a CER of 4.46\%.
\end{abstract}
\end{frontmatter}
\setcode{utf8}
\section{introduction}
\begin{figure*}[]

\frame{\includegraphics[width=1.1\textwidth]{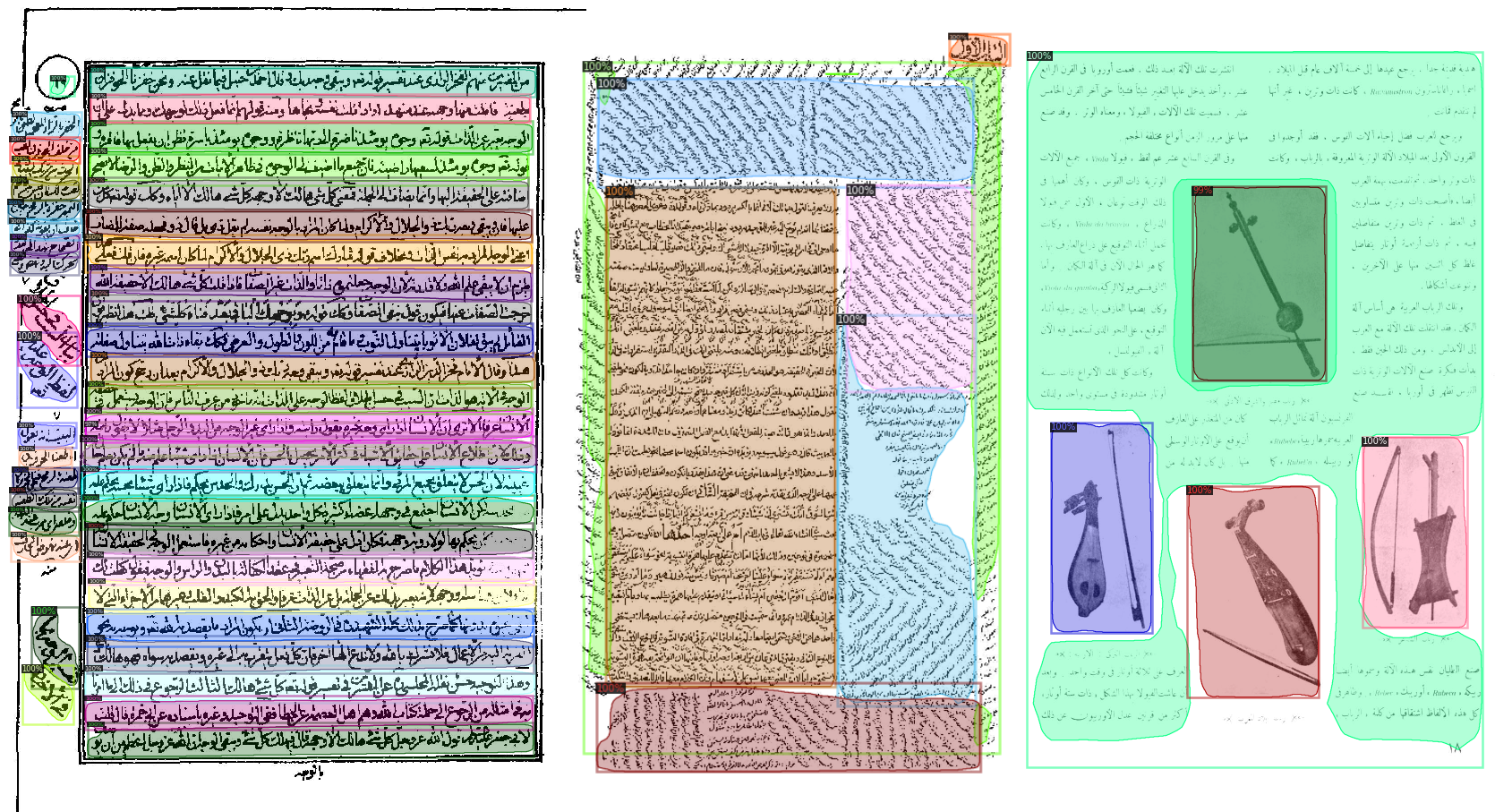}}

\caption{Example Segmented Page/Text Line images of historical printings}
\label{figure_3}
\end{figure*}

\begin{figure*}[]

\frame{\includegraphics[width=1.1\textwidth]{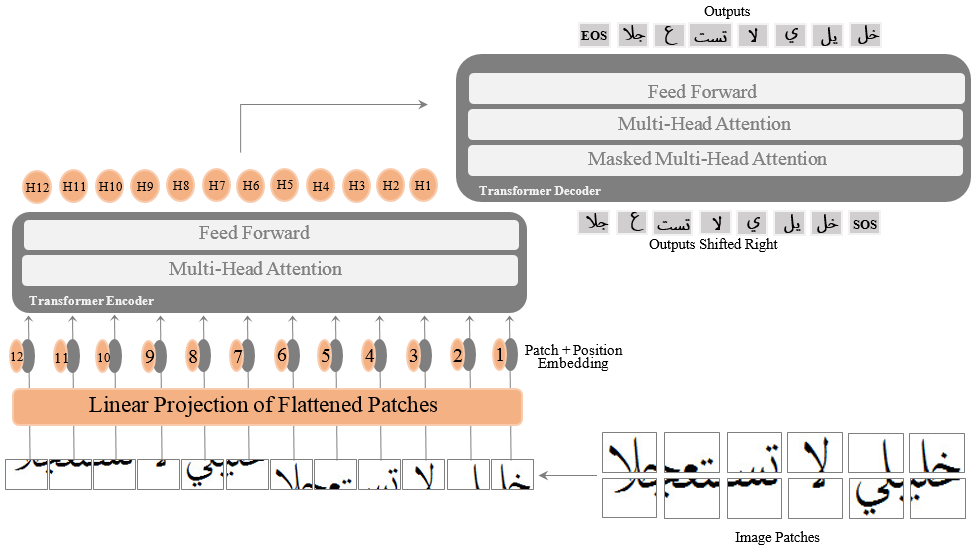}}

\caption{Proposed Architecture, End-to-End Architecture}
\label{arch}
\end{figure*}

Arabic is spoken by over 433 million throughout the world and is an official language in 26 countries \citep{theeditorsofencyclopaedia}. Arabic writing is an essential mode of communication. Humans created new strategies that are not only effective but also rapid as technology advanced. No one can dispute that electronic media has supplanted paper in recent years. Electronic equipment is used to copy, scan, send, and save documents. The appeal of these devices arises from the fact that they enable future data recovery to be simple and rapid. The current upsurge of interest in Egypt's cultural legacy as represented by its literary sources has led to studies and strategies to solve the challenge of making historical complete texts available: How can we significantly reduce the cost of transforming scanned page pictures into searchable full text? (both in terms of time and money). Adaptations of current tools in the field of Optical Character Recognition (OCR) \citep{gruuening2017robust, kahle2017transkribus, neudecker2019ocr} in response to these issues, efforts have been undertaken to convert scanned or printed text pictures, as well as handwritten text, into editable text for further processing. This method enables computers to identify text on their own. It's like a hybrid of human sight and intellect. Although the eye can perceive the text in the pictures, it is the brain's responsibility to analyze and comprehend the extracted information. Several challenges may occur during the development of a computerized OCR system. There is relatively little discernible difference between some letters and numerals for computers to interpret. For example, the computer might have trouble distinguishing between the digit "\<“١”>" -one in Arabic- and the letter "\<“أ”>" - Alef letter-. In the following ways, images in ancient literature differ from those on current book pages: Most of these flaws can be found in historical fonts, including historical spelling variants, identical words spelled differently not only between books of the same period but even within the same book, slightly displaced characters (due to historical printing processes), cursive letters, fuzzy character boundaries due to ink creep into the paper over time, paper degradation resulting in dark backgrounds, blotches, cracks, dirt, and bleed-through from the next page. \citep{stahlberg2016qatip, darwish2020enhanced, clausner2018icfhr}. Reader's Digest built the first commercial system that used OCR to input sales reports into a computer in 1955 \citep{herbert1982history}, and since then, OCR technology has proven to be incredibly beneficial in computerising physical office paperwork \citep{CaereCorporationHistory}. Text recognition research has traditionally focused on Latin characters, such as English, with non-Latin scripts, such as Arabic, just being examined in the last two decades \citep{lawgali2015survey}. While OCR technology has advanced in recent years, it still falls short of the accuracy necessary for historic Arabic printings \citep{althobaiti2017survey}. This is due to the use of images, aesthetic border elements and decorations, and marginal remarks in the arrangement as shown in  \autoref{page_seg}, \autoref{figure_3}. Text and non-text segmentation cannot yet be completely automated with high accuracy \citep{lawgali2015survey}. Furthermore, non-standardized fonts present a considerable hurdle to OCR algorithms\citep{althobaiti2017survey}.\newline
Long Short Term Memory Recurrent Neural Networks (LSTM) \citep{hochreiter1997long, graves2013generating} trained using a Connectionist Temporal Classification (CTC) \citep{graves2006connectionist} decoder specialized for OCRs, Attention mechanisms \citep{bahdanau2014neural}, Self-Attention \citep{vaswani2017attention}, Transformers, and end-to-end architectures \citep{wang2021large, baevski2020wav2vec} were recently introduced as a significant milestone. These milestones have improved the recognition process, increasing both text and character recognition accuracy.\newline Pre-processing \citep{alginahi2010preprocessing, bui2017selecting, bieniecki2007image}, segmentation \citep{lee2019page, ayesh2017robust}, feature extraction, classification, and post-processing \citep{khirbat2017ocr, bassil2012ocr, boiangiu2009ocr} are the five major stages of OCR system development. At each step, different tactics are employed. \newline The major contributions of this research can be summarized as follows:
\begin{enumerate}
    \item We generated the largest dataset for Arabic OCR, with 30.5 million images (that is, text lines) and 270 million words associated with their text ground truth, including diverse fonts and writing styles. As shown in \autoref{tbl:dataset}, the APTI Dataset - currently the largest amongst those consisting of Arabic text lines/sentences - includes only (45 million) words.
    \item We present a unique transformer-based architecture (\autoref{arch}) for Arabic OCR that is end-to-end by employing a transformer encoder as a feature extractor rather than the traditional CNN models.
    \item We proposed and developed a complete OCR for Arabic Handwritten text line pipeline that comprises all processes from taking an image as input to applying pre-processing, Page/Text line segmentation, image enhancement, text line image to text transcription, and finally using post-correction approaches to increase recognition performance. Up to our knowledge, no other studies in the literature present a complete pipeline of an Arabic OCR.
\end{enumerate}

The rest of this paper is organized as follows: \autoref{sec2} discusses the related work, \autoref{sec3} presents the Methods and Materials (including the constructed dataset), \autoref{sec4} presents the Results, Analysis, and Discussion, and \autoref{sec5} concludes this work while highlighting its limitations with some recommendations for future work.

\section{Related Work: A Literature Review of Arabic OCR Approaches}\label{sec2}
This section discusses previous research and applications addressing Handwritten text recognition challenges in Arabic, as well as the methodologies, datasets, strengths employed, and drawbacks.

Neural Networks-based algorithms have traditionally outperformed conventional machine learning techniques (Support Vector Machines for instance) when constructing Arabic OCRs. In 2004, \citep{haraty2004arabic} Their approach is made up of three primary components. Binarization, skeletonization, and character block extraction are used in a heuristic technique to extract picture information. The extracted block and character classification are then evaluated using a combination of two Neural Network architectures. They trained their architecture using a dataset of 10,027 samples and tested it on 2132 samples collected from students around the Lebanese American University, achieving a 73\% accuracy rate in character recognition.

\citep{dreuw2009writer} Developed an OCR that utilising Maximum Mutual Information (MMI) and Minimum Phone Error (MPE). They also utilised a neural network to extract features. They claimed that the proposed methods can distinguish between handwritten and machine-printed scripts. Their experiment, which was carried out on the IFN/ENIT Arabic handwriting database, resulted in a 50\% reduction in word-error rate.

To recognise Arabic characters, \citep{addakiri2012line} demonstrated a neural network-based online handwriting system The proposed system's three main components are preprocessing, feature extraction, and classification. All characters are preprocessed at the initial step to increase their visual quality. The image of each character is transformed to a 2-bit image (binary image). A backpropagation technique was then used to train the neural network. Finally, Neural Networks are employed in the detection of Arabic characters. When evaluated on 1400 writing styles, this method has an accuracy rate of 83\%.

\citep{osman2020efficient} present an Arabic OCR pipeline that takes as input a scanned image of the Arabic Naskh script and apply Pre-processing techniques, Word-level Feature Extraction, Character Segmentation, Character Recognition, and Post-processing. This paper also employs word and line segmentation. Finally, a neural network model for character recognition is proposed in the study. The system was evaluated on a variety of accessible Arabic corpora datasets(watan 2004 and subset of APTI), with an average character segmentation accuracy of 98.66\% percent, character recognition accuracy of 99.89\% percent, and total system accuracy of 97.94 percent.

On the KHATT dataset, \citep{ahmad2020deep} A Deep Learning benchmark was reported. On the images, they employed pre-processing and image augmentation. The pre-processing stage consists of removing white extra spaces and de-skewing skewed text lines. They employ a network that incorporates Multi-Dimensional Long Short-Term Memory (MDLSTM) and Connectionist Temporal Classification (CTC). According to them, MDLSTM has the advantage of scanning Arabic text lines in all directions (horizontal and vertical) to cover dots, diacritics, strokes, and tiny details. They earned an 80.0\% Character Recognition rate.

\citep{fasha2020hybrid} developed a model for recognising Arabic printed text without character segmentation using a hybrid DL network They put the classifier to the test with a custom dataset of over two million word samples generated by (18) different Arabic font variations. The proposed model employs a Convolutional neural network and a Recurrent neural network. These networks are linked end-to-end to conduct word-level recognition without character-level segmentation.

\paragraph{ Hijja, a dataset of Arabic letters produced entirely by children aged 7–12, was proposed by}\citep{altwaijry2021arabic}. They trained convolutional neural networks on the proposed dataset as well as the Arabic Handwritten Character Dataset, yielding accuracies of 97\% and 88\%, respectively.\newline
According to the findings of this survey. Many issues have been identified. To begin, several approaches for Arabic handwritten text recognition are ineffective for handwritten fonts. In addition, a lot of researchs focuses solely on the recognition phase. There aren't many solutions that construct a complete pipeline from segmentation to post-processing. While there has been a lot of study into developing Arabic OCR for general use, there has been very little research into trying with Arabic handwritten recognition. Furthermore, substantial and diversified datasets for Arabic handwritten recognition are scarce. As a result, constructing large Deep Learning models to aid in the problem of OCR of Arabic handwritten fonts has limitations. The purpose of this work is to overcome previous limitations by employing pre- and post-processing techniques, assessing state-of-the-art Deep Learning models, and training those models on large and diverse datasets.

\section{The Proposed Methodology: Constructing a Complete Pipeline}\label{sec3}

In this section, the essential components of the proposed pipeline are presented, beginning with creating the largest Arabic dataset to overcome the lack of ground truth for the great majority of ancient Arabic manuscripts (in \autoref{sec:dataset}). Followed by (in \autoref{sec:dataaug}) augmentation techniques that generate additional samples of the Arabic sentences resembling - for instance - marginal notes and rotated lines-of-text that can be found throughout historical manuscripts. Within pages, text paragraphs and marginal notes are then segmented - while discarding spaces and illustrations - before segmenting the text-lines. This segmentation component is detailed in \autoref{sec:seg}. To further improve the final results, four different Image enhancement techniques are applied sequentially to each segmented text-line(\autoref{sec:enhance}), namely, contrast enhancement, edge detection, text locating, and finally a median filter for noise removal. To achieve an accurate transcription of handwritten historical Arabic documents (including an efficient identification of the Arabic diacritics) an innovative OCR is developed by utilising Transformers (detailed in \autoref{sec:model}). Finally, the last phase of the proposed pipeline includes two post-correction methods (\autoref{sec:post_corr}), employed to decrease the character error rate, and thus ensuring a better overall performance.


\subsection{Dataset Collection}\label{sec:dataset}

\begin{table}[htbp]

\begin{center}

\caption{Selected Arabic text datasets.
\\ \scriptsize (N/A) indicates that information is not available}
\label{tbl:dataset}
\small

\resizebox{11cm}{!}{%
\begin{tabular}{@{}l l   c  c  c c c     @{}}
\toprule
 & Dataset & \#Words & \#Characters & \#Fonts &\#Font Size & \#Font Styles \\\midrule
\multirow{7}{*}{} & APTI(words)
 &  45,313,600  & 648,280 &  10 & 10 & 4     \\
 & IFN/ENIT(words) & 26459 &  212,211 & 1 (Handwritten) & N/A & 1  \\
 &  HACDB(characters) & N/A & 6600 & 1 (Handwritten) & N/A & 1  \\
 & APTID / MF(character,text) & N/A &  27,402 & 10 & 4 & 10      \\
 &  KHATT(text) & 400 & 7900 & 1 (Handwritten) & N/A & 1      \\
 & \textbf{Proposed Dataset} & \textbf{270m} & \textbf{1.6 billion} & \textbf{12} & \textbf{13} & \textbf{13}  \\
 \bottomrule
\end{tabular}%
}
\normalsize
\end{center}
\end{table}

A major contribution of our work is to create the largest dataset for Arabic OCR, we used the same dataset that was created in the first work \citep{mostafa2021ocformer}. The dataset consists of images with Arabic text that were obtained from the web \citep{yousef2019learning}, together with their ground truth. Arabic diacritics are used in a section of the text. Furthermore, we employed a variety of Arabic fonts that closely resembled archaic fonts used in historical printings dating back to the $18^{th}$ century. (Not simply an image, but a page with 15 lines) There are four categories of images: Full sequences, or visuals with more than five words, Short sequences, or visuals with five or less syllables, Full sequences with diacritics, which are images with more than five diacritic words, and Short sequences with diacritics, which are images with five or less diacritic words. In addition, we gathered the handwritten printings from the KHATT database \citep{mahmoud2014khatt}, which has unrestricted handwritten Arabic Texts produced by 1000 distinct authors. This distinction has been made so that the model may be trained on all sorts of sequences and sentence positions that may occur in historical printings. For example, marginal notes are Short sequences, but artistic borders are Full sequences, guaranteeing that the model can train on all sorts of texts and help in the segmentation stage for printings and handwritten text. The proposed dataset is a comprehensive, multi-font, multi-style Arabic text recognition dataset. The dataset was built with a range of characteristics to ensure the diversity of the writing styles. This comprises the use of different fonts, styles, and noise patterns to the characters used to make the pictures. The database was created using 12 Arabic fonts and a range of font styles. There are 30.5 million single-line pictures, nearly 270 million words, and 1.6 billion characters in the collection. The ground truth, style, and typography of each image are all available. Because it is well recognised that too little training data results in poor generalization for Deep Learning models \citep{lecun1989generalization} as in the previous works, we attempted to tackle this problem in our approach by creating a large sample size dataset, the statistics of different datasets employed in Arabic OCR \citep{slimane2009database, pechwitz2002ifn, lawgali2013hacdb, pechwitz2002ifn, mahmoud2014khatt} and our proposed dataset are shown in \autoref{tbl:dataset}.

\begin{figure*}[]

\frame{\includegraphics[width=0.9\textwidth]{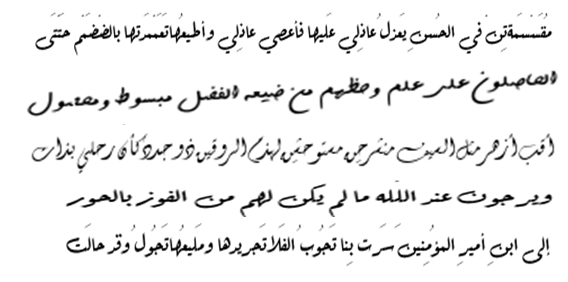}}

\caption{Sample Line of proposed dataset}

\end{figure*}

\subsection{Image Augmentation}\label{sec:dataaug}

Similarly to the previous study, we augmented the dataset to boost variation and noise. We employed cropping, padding, horizontal flipping, zooming, and rotating the image to ensure the model's resilience while introducing as much noise as possible. Shearing and altering the brightness are two often utilised augmentation strategies when training big neural networks. Furthermore, we used Random Angle Rotation and Line Stretching to mimic the writing arrangement patterns present in the bulk of historical manuscripts. Marginal notes, for example, are frequently written in white spaces on a page sideways.

\subsection{Segmentation}\label{sec:seg}

We divide the images into regions of interest. On a small sample of the hand-annotated images, we trained a pre-trained Mask-RCNN, Detectron-2 \citep{wu2019detectron2}. Detectron-2 is a complete overhaul of Detectron that began with the masks CNN-benchmark. Detectron-2 is adaptable and scalable, with the ability to train quickly on single or multiple GPU servers. Detectron-2 provides high-quality implementations of cutting-edge object identification methods such as DensePose, panoptic feature pyramid networks, and several variants of Facebook AI Research's pioneering Mask R-CNN model family (FAIR). Mask- RCNN is a cutting-edge model for instance segmentation that was built on top of Faster R-CNN, a region-based convolutional neural network. It produces bounding boxes for each object as well as its class label, together with a confidence score. We suggest a method for segmenting the page and line effectively. We used two types of segmentation, \textbf{Page Segmentation}, and \textbf{Text Line Segmentation}.

\begin{figure*}[]

\frame{\includegraphics[width=1.1\textwidth]{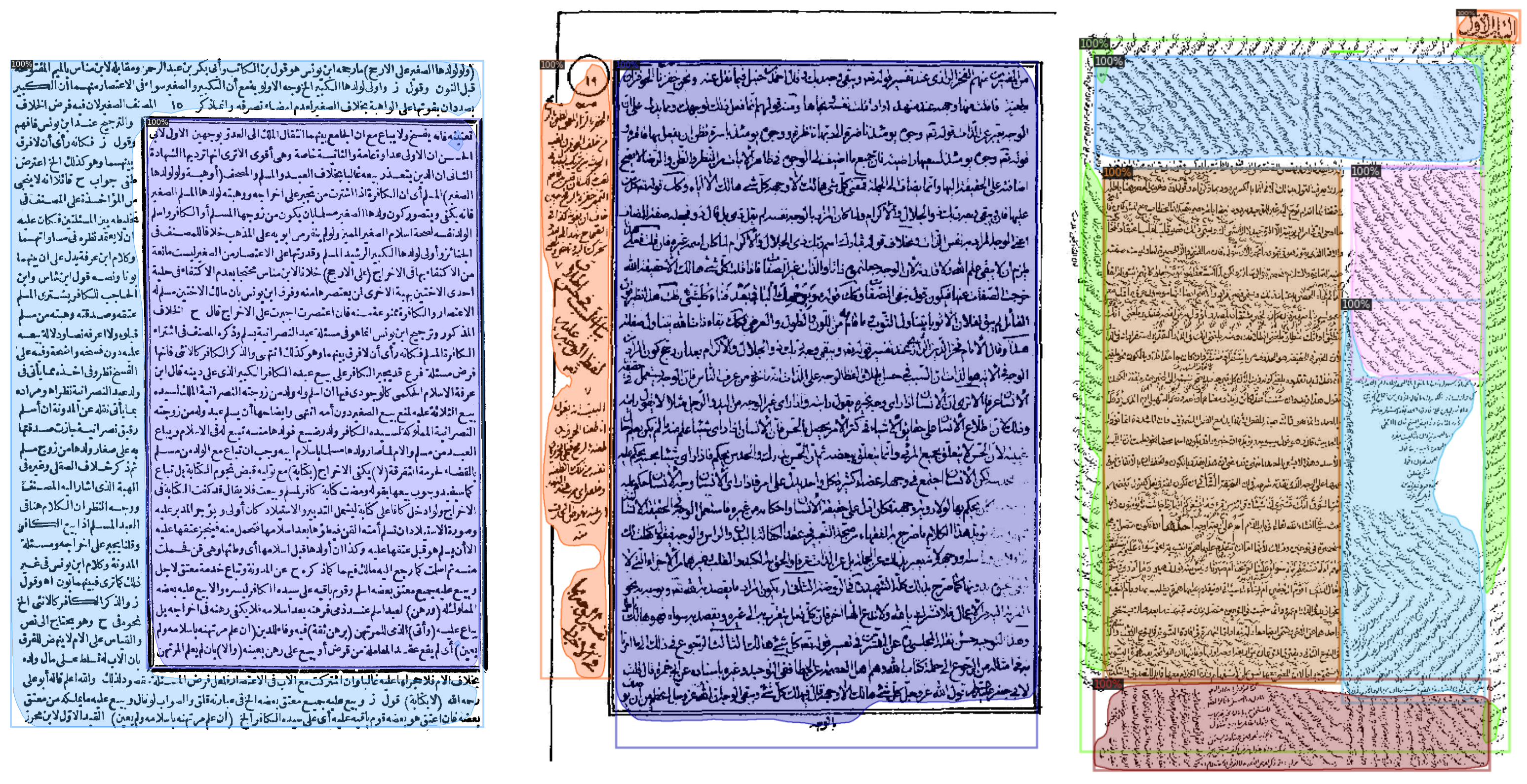}}

\caption{Dataset's sample images of Segmented historical printings}
\label{page_seg}
\end{figure*}

\subsubsection{\textbf{Page Segmentation}}

\paragraph{Page segmentation} \citep{kise2014page} is the process of extracting homogeneous components from page images. As components, text blocks or zones, text-lines, graphics, tables, and images are widely employed. Component classification is part of the page segmentation job, in which the model identifies each component as a text block, graph, or marginal notes. It is crucial to recognise that these functions are not always separate; they are sometimes viewed as two sides of the same coin. In fact, the work of page segmentation and classification is frequently referred to as "(physical) layout analysis." Some approaches, on the other hand, are intended to work without classification. We trained Detectron-2 on a sample of handcrafted annotations, containing overlapped text such as marginal notes which are common in historical printings, initialized with three classes:
\begin{enumerate}
    \item  Text Block
    \item Graphs or picture
    \item Marginal Notes
\end{enumerate}

Note that we did not have a table component in our dataset, so we did not specify a class in the Page Segmentation process. Sample of page segmentation is illustrated in \autoref{page_seg}.

\subsubsection{\textbf{Text Line Segmentation}}
\paragraph{Text line segmentation} \citep{barakat2018text, younes2015segmentation} is a critical pre-processing step in document analysis that is particularly tough for handwritten material. Text lines have historically been important for assessing document layout, determining the skew or orientation of a page, and indexing/retrieval based on word and character recognition. Although machine-printed text line segmentation is a solved problem, freestyle handwritten text lines remain a substantial challenge. This is due to the fact that handwritten text lines are frequently curved, have nonuniform space between lines, and may have spatial envelopes that overlap. Handwritten document analysis is particularly complicated by irregular layout, varied character sizes resulting from diverse writing styles, the presence of touching lines, and the lack of a well-defined baseline. The existence of diacritical components complicates the task even more in Arabic. Detectron-2 was trained using a sample of handcrafted annotations incorporating numerous anomalies in writing styles, similar to page segmentation. The total loss of text line segmentation is 0.1114.

 
\begin{figure}
\centering

\frame{\includegraphics[width=0.80\textwidth]{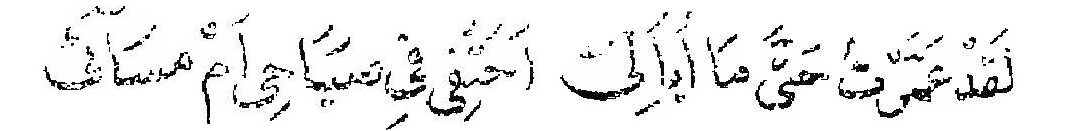}}

\vskip 0.5mm
\centering

\frame{\includegraphics[width=0.80\textwidth]{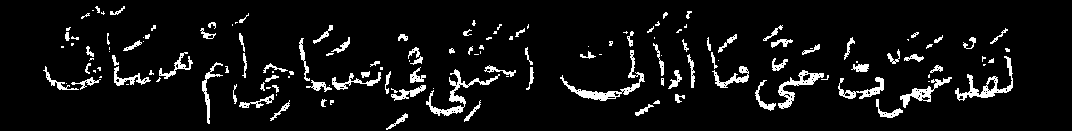}}

\vskip 0.5mm
\centering

\frame{\includegraphics[width=0.80\textwidth]{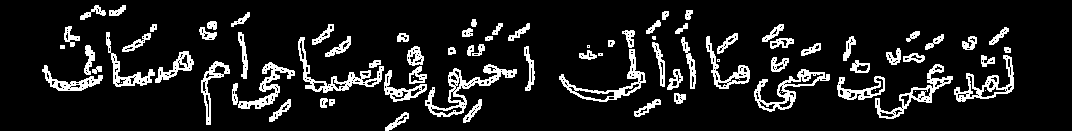}}

\vskip 0.5mm
\centering

\frame{\includegraphics[width=0.80\textwidth]{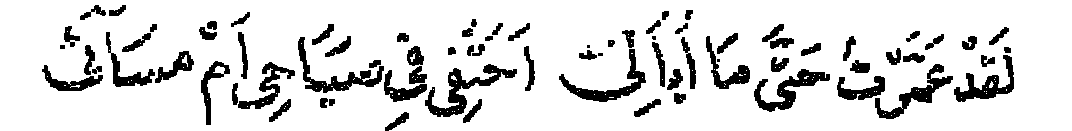}}

\vskip 0.5mm
\centering

\frame{\includegraphics[width=0.80\textwidth]{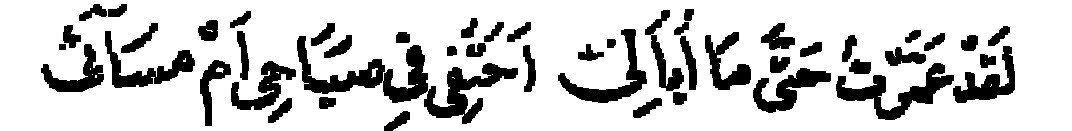}}

\vskip 0.5mm
\centering

\frame{\includegraphics[width=0.80\textwidth]{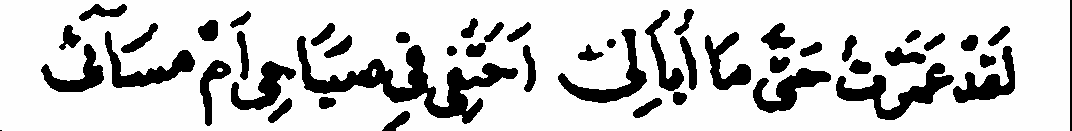}}

\caption{Example of Applied Enhancement Process On Proposed Dataset}
\label{enhanced}

\end{figure}

\subsection{Image Enhancement}\label{sec:enhance}

 In this work, we aim to show the effectiveness of improving the input images, helping the model extract valuable features that would be otherwise missing. Our suggested technique works by enhancing the contrast of scanned documents and then constructing an edge map from the contrast-enhanced image to locate text regions. We use the text position information to apply a median filter to remove noise, similar to the salt and pepper effect.

We adopted three enhancing phases from \citep{chen2012efficient}, tackling the problem of improving the image quality to improve the recognition process.
\begin{itemize}

\item In the first phase, \textbf{Contrast Enhancement}, improves the contrast of the original text image to increase the luminosity difference between the text and the backdrop.\newline

\item In the second phase, \textbf{Edge Detection}, They employ the Sobel edge detection approach to generate an edge picture that represents the text portion of the source image. To detect distinct directions, four edge pictures are created using four different masks. Following the generation of the four edge pictures, the detection result is built by computing the average output. According to a predefined threshold, the detection result is converted into a binary image.

\item During the third stage, \textbf{Text locating}, By initially locating the text, you may create a background-like image of the original text picture. After locating the text pixels, utilise interpolation to replace them with new ones. Images CEI and I were utilised to find the text pixels in I, and $EI bin$ is required. First, using the established threshold $th c$, CEI is translated into its binary counterpart, $CEI bin$. The text location image, TLI, is then created by combining $CEI bin$ and $EI bin$.

\end{itemize}

Finally, we utilised a \textbf{Median Filter}, a non-linear digital filtering technique commonly used to remove noise from an image or signal for certain noise types such as "Gaussian," "random," and "salt and pepper." The median filter substitutes the centre pixel of a M × M neighborhood with the window's median value. It is worth noting that noise pixels are regarded to be separate from the median. Following on this notion, a median filter may reduce this kind of noise problem. This filter is used after text localization to diminish noise pixels in text-line pictures. The enhanced process on the propsed dataset is shown in \autoref{enhanced}.

\subsection{\textbf{Model Architecture}}\label{sec:model}

The transformer \citep{vaswani2017attention} is a deep learning model that assumes the attention process, weighting the importance of each component of the input independently. The transformer has been a discovery in NLP since its conception. Transformers enable training on bigger datasets than was previously feasible, prompting researchers to create pre-trained models like BERT \citep{devlin2018BERT} (Bidirectional Encoder Representations from Transformers) and GPT (Generative Pre-trained Transformer), which are trained on enormous language datasets.

The training pipeline goes as follows: The Enhanced Text lines images pass through the image Transformer as an encoder for feature extraction. Then, we initialize the vanilla Transformer model with two encoders to capture a representation of the image, two decoders to construct the character piece sequence while accounting for the encoder output and preceding generation, one attention head, and 128 hidden dimensions. Finally, a cross-entropy loss function with Label Smoothing.

\subsubsection{\textbf{Encoder}}

BEiT, which stands for Bidirectional Encoder representation from Image Transformers, was utilised as the encoder. The authors propose a masked image modelling challenge to pre-train vision Transformers based on BERT, which is well-known in the field of natural language processing. Each image in our pre-training contains two views: image patches (such as 16x16 pixels) and visual tokens (i.e., discrete tokens). They advocated first "tokenizing" the original picture into visual tokens. Then, using the backbone Transformer, mask several image patches at random. The pre-purpose training's is to recover the original visual tokens from corrupted picture patches. After pre-training BEiT, directly fine-tune the model parameters on downstream tasks by adding task layers to the pre-trained encoder. According to experimental data, the BEiT model surpasses previous pre-training approaches in image classification and semantic segmentation. Base-size BEiT, for example, obtains 83.2 percent top-1 accuracy on ImageNet-1K.

\subsubsection{\textbf{Decoder}}
A self-attention mechanism, an attention mechanism over the encodings, and a feed-forward neural network comprise the decoder. In addition to the two sub-layers in each encoder layer, the decoder injects a third sub-layer that conducts multi-head attention on the encoder outputs. Residual connections, similar to the encoder, are used around each sub-layer, followed by layer normalisation. Then, in the decoding stack, change the self-attention sub-layer to prohibit locations from attending to following counterparts. Furthermore, the first decoder receives positional information and embeddings of the output sequence as input rather than encodings. The transformer must not forecast using the present or future output, which is why the output sequence is partially masked to prevent this reverse information flow. To obtain the output probabilities across the vocabulary, the final decoder is followed by a final linear transformation and softmax layer.


\subsection{\textbf{Post-Correction}}\label{sec:post_corr}
\subsubsection{\textbf{Word Beam Search Decoding Algorithm With Dictionary
}}

\begin{figure*}[]
\begin{center}
    
\frame{\includegraphics[width=0.4\textwidth]{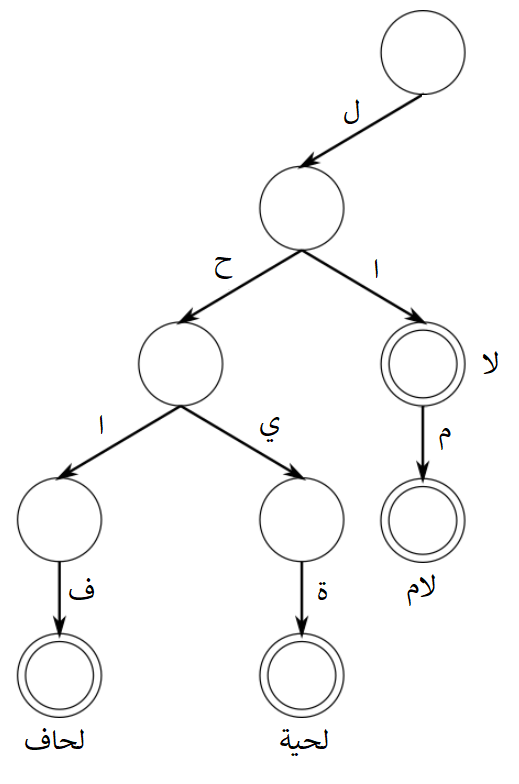}}

\caption{Example of Prefix Tree Between Two Words}
\label{post_corr}
\end{center}

\end{figure*}
\setcode{utf8}
The transformer decoder outputs the characters purely depending on the image. In the case of poor dotting of the Arabic letters in the manuscripts, the decoder for example might mistake between the letter Yaa' "\<“يـ”>" and the letter baa' "\<“بـ”>" at the beginning of a word. Also, some of the Arabic letters do not connect with the following letter, which might cause the model to falsely think that there is a white space between the two letters, for example, the word "\<“جمال”>" the decoder might output
"\<“جما ل”>". There were various decoding solutions to try to overcome this problem, Beam search decoding, Beam search decoding with a character language model, token passing (word language model), and Word beam search (WBS) \citep{scheidl2018word}, which is a combination between beam search and token passing. We chose the WBS to be our decoding approach, but it was mainly proposed on RNN decoder in a sequence to sequence model, so we had to adapt the algorithm to work with the transformers decoder. The WBS has two modes, word mode and non-word mode. The WBS first needs an Arabic dictionary to create its Prefix Tree from it, which is a tree that the model navigates through when it starts a new word in the decoding process. At the beginning of the decoding process, it starts in the non-word mode where nothing new happens yet, the decoder starts decoding letters, numbers, or punctuation marks, once it decodes a letter it switches to the word mode where it can't decode numbers or punctuations until it completes a word from the prefix tree, To understand it well, let's assume the model started decoding in non-word mode and decoded the letter laam "\<“ل”>", it switches to word mode, and it goes to the prefix tree that was generated before this using a dictionary. to simplify we created a prefix tree having only 4 possible words that start with the letter Laam "\<“ل”>" shown in \autoref{post_corr}. the model then is forced to decode based on this tree by looking at the possible next letters which in this case the letter Alif "\<“ا”>" and Haa' "\<“ح”>" and only choosing from these letters, and only if it completes a word, it switches back to non-word mode. To be able to correct the misspelled characters in high accuracy, we build a huge dictionary using King Saud University Corpus of Classical Arabic(KSUCCA) made up of Classical
Arabic texts dating between the 7th and early 11th century\citep{alrabiah2014king} which consist of 202 0063 Sentences and 46 million words, 934 177 of them are unique.

\subsubsection{\textbf{Auto-correct Using BERT Model}}

BERT (Bidirectional Encoder Representations from Transformers) \citep{devlin2018BERT} has created a big impact in the Machine Learning field by showing cutting-edge findings across a wide range of NLP tasks such as Question Answering, Natural Language Inference, and others. The main technological breakthrough of BERT is the use of bidirectional training of Transformer, a prominent attention model, to language modelling. This is in contrast to prior studies, which looked at a text sequence from left to right or a combination of left to right and right to left training.

To rectify words mistakenly transcribed from the OCR output by Masked Language Modeling (MLM), a language task common in Transformer systems nowadays, we apply a pre-trained BERT model. It entails masking a portion of the input and then training a model to predict the missing tokens, thus reconstructing the non-masked input. MLM is frequently used in pre-training jobs to teach models textual patterns from unlabeled data. The first step is how to know if a word in the sequence is misspelled. BERT tokenizer breaks the text into word pieces that are in its vocab. Thus, if a word is broken into small pieces (shown with \#), then it is misspelled. We use this fact to detect errors.

\begin{figure*}[]
\begin{center}

\includegraphics[width=13cm, height=7cm]{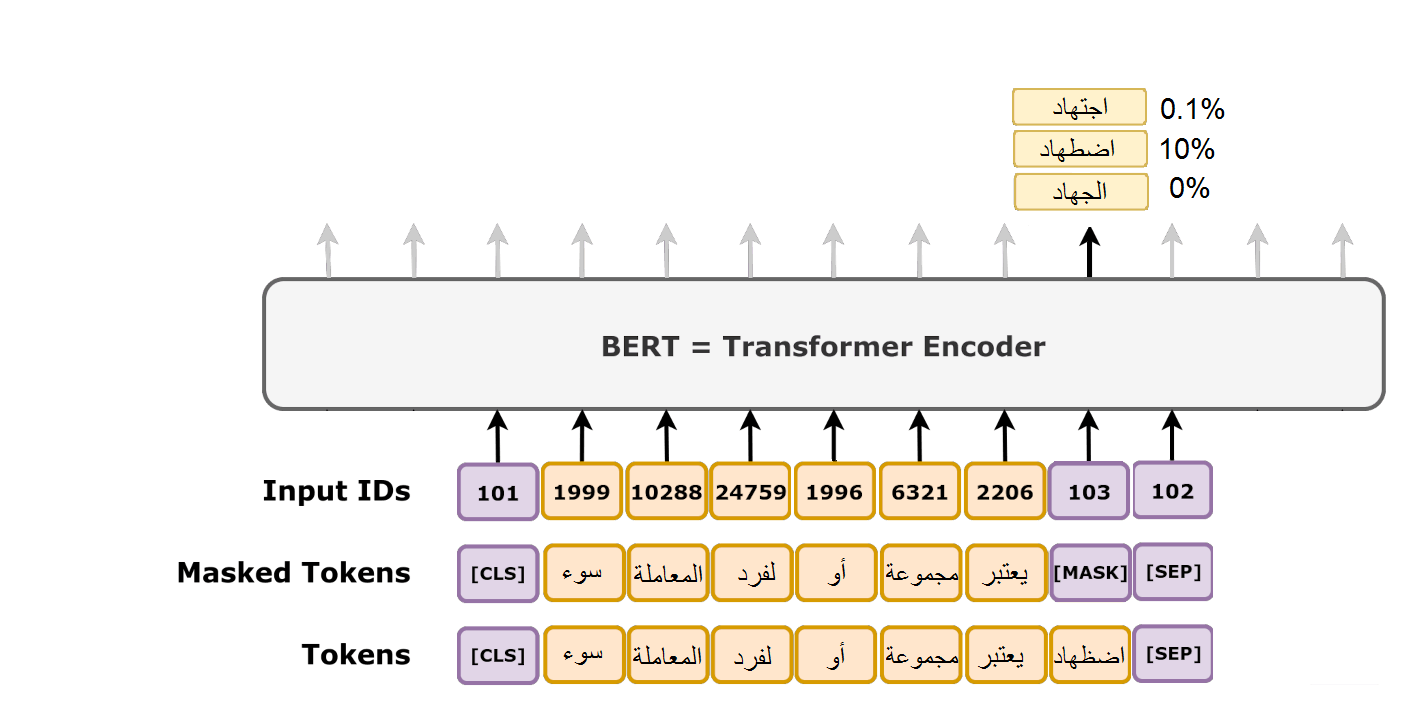}

\caption{Auto-Correct Using BERT MLM}
\label{post_corrBERT}
\end{center}

\end{figure*}

As a result, each wrong word is substituted by a "[MASK]" token in order to produce a BERT forecast. Following this stage, we have the embedding for each word in the input phrase, also known as input embedding. Following that, it takes the embedding sequence as input, looks for [MASK] tokens in the input, and then attempts to estimate the original value of the masked words based on the context supplied by the non-masked words in the sequence. BERT also takes segment embedding, which is a vector used to differentiate numerous phrases and aid in word prediction. For example the segment vector for "\<“أحمد ذهب إلى المتجر. و اشترى زجاجتين من الحليب.”>", would be [0, 0, 0, 0, 0, 1, 1, 1, 1, 1, 1]. The model outputs the probabilities of error corrections for each “[MASK]” token based on its context, however, we don’t take all the predicted words, only take words with two edit distances away from the original misspelled word. Cross-entropy is used to determine the loss, which measures the relative entropy between two probability distributions over the same collection of occurrences. To compute cross-entropy between P and Q, you intuitively derive entropy for Q using probability weights from P. As the anticipated probability diverges from the actual label, cross-entropy loss grows. The cross-entropy between two probability distributions, such as Q from P, may be expressed formally as H(P, Q), where H() is the cross-entropy function, P may be the target distribution, and Q is the target distribution's approximation.

\begin{algorithm}[t]
\caption{Masked Language Model Post-Correction}\label{alg:cap}
\vskip 1mm
 \hspace*{\algorithmicindent} \textbf{Input:} \text{Input sentence "S"} \\
 \hspace*{\algorithmicindent} \textbf{Output: }\text{Corrected sentence}
     \algrule

\begin{algorithmic}

\For{i =1 to \text{T}}
{

\State Masked Sentence $\gets$ \text{MaskMisspelledWord(S)}
\vskip 0.5mm
\State Output $\gets$ \text{BERTModel(Masked Sentence)}
\vskip 0.5mm
\State P $\gets$ \text{CrossEntropyScore(Output)} \Comment{Syntatic Structure}
\vskip 0.5mm
\State PS $\gets$ \text{PerplexityScore(Output)} \Comment{Sematic Structure}  
\vskip 0.5mm
\State S $\gets$ \text{MultiplicationScore(PS,P) }

\EndFor 

}
\State \textbf{Return} S

\end{algorithmic}
\label{alg:MYALG}
\end{algorithm}

\begin{equation}
    H(P,Q) = \sum_{x}^{X} P(x) * log(Q(x))
\end{equation}

  In our approach, P is the output prediction tensor from the BERT model, and Q is each sentence with the replaced predicted word. Then, we evaluate the predicted sentences using the perplexity score, an evaluation metric for language models to measure the sentence structure. The final step is to multiply each predicted sentence BERT perplexity score shown in \autoref{equation} with its cross-entropy score to get the final output sentence. \autoref{post_corrBERT} shows an illustration of our approach pipeline. The number of misspelled words in a sentence and how many words are predicted for each misspelled word affect computational time. Algorithm.~\ref{alg:MYALG} shows all steps of Auto-correct using BERT model.

\begin{equation}\label{equation}
    P(W) = P(w_1)P(w_2\mid w_1)P(w_3\mid w2,w1)...P(w_N\mid w_{N-1},w_{N-2})
\end{equation}

\section{Discussion}\label{sec4}

\subsection{ConvNet}
\paragraph{In our last work} we used ResNet101 as our feature extractor. From our experiments, ResNet101's layers didn't impact the model's performance and provided the same results as its smaller variant ResNet18. This indicates that there was no meaningful information was extracted by the additional layers in the ResNet101.In addition, ResNet101 takes longer training, due to resource scarcity, which was not feasible. Then we implemented EfficientNet in conjunction with Noisy Student Training. EfficientNet is a scaling architecture based on convolutional neural networks. EfficientNet scales all depth/width/resolution dimensions evenly using a compounded coefficient. Unlike traditional technique, which varies these elements arbitrarily, the EfficientNet scaling approach reliably enhances network breadth, depth, and resolution with a set of predefined scaling coefficients. Noisy Student Training is a semi-supervised learning approach that works effectively even with a large amount of labelled data. Noisy Student Training extends the concept of self-training and distillation by using equal-or-larger student models and noise introduced to the learner during learning. Despite EfficientNet with Noisy Student Training lowered training time and increased model performance, it falls short of the cutting-edge Vision Transformers. The comparison between different backbones is illustrated in \autoref{tbl:downstream_results}

\begin{table}[htbp]

\begin{center}
\small
\caption{Comparasion Between Different Backbones}
\label{tbl:downstream_results}
\resizebox{12cm}{!}{%

\begin{tabular}{@{}l l  c  c  c  c  c  c @{}}
\toprule
Backbone & LR-Scheduler & Encoder/Decoder & Hidden Units & Number of Heads  & CER \\\midrule

\multirow{2}{*}{ResNet-18} & LR-Scheduler & 4 & 256 & 4 & 8.52 \\
 & ONE CYCLE-LR & 2 & 128 & 1 & 8.21\\\midrule

\multirow{2}{*}{ResNet-101} & STEP-LR 
 & 4 & 256 & 4 & 7.87 \\
 & ONE CYCLE-LR & 2 & 128 & 1 & 7.42 
\\\midrule

\multirow{1}{*}{EfficientNet-V2} & ONE CYCLE-LR & 2 & 128 & 1 & 6.89 \\\midrule

\multirow{1}{*}{NFNet-l0} & ONE CYCLE-LR & 2 & 128 & 1 & 6.11 \\\midrule

\multirow{1}{*}{EfficientNet-B4-ns} & ONE CYCLE-LR 
 & 2 & 128 & 1 & 5.45 \\\midrule

\multirow{1}{*}{EfficientNet-B2-ns} & ONE CYCLE-LR
 & 2 & 128 & 1 & 5.27  \\\midrule

\multirow{1}{*}{\textbf{BEIT$_B$(End-to-End)}} & \textbf{ONE CYCLE-LR} 
 & \textbf{4} & \textbf{256} & \textbf{4} & \textbf{4.64} \\

 \bottomrule
\end{tabular}%
}
\normalsize
\end{center}
\end{table}

\subsection{Optimization}

The lack of resources forced us to find efficient implementations since training from scratch takes an immense amount of time and computational power. To remedy this, we tested popular optimized transformer models. We started with Linformer \citep{wang2020linformer}, a linear Transformer that employs a linear self-attention mechanism to address the self-attention bottleneck in Transformer models. Through linear projections, the initial scaled dot-product attention is divided into several smaller attentions, resulting in a low-rank factorization of the original attention. In terms of both space and temporal complexity, it lowers self-attention to an O(n) operation. Then we attempted Performers \citep{choromanski2020rethinking}, which employ the Fast Attention Via Positive Orthogonal Random Features (FAVOR+) mechanism, utilising softmax and Gaussian kernel approximation approaches.

Performers are the first linear designs that are perfectly compatible with regular Transformers (by little fine-tuning), providing clear theoretical guarantees like as unbiased or nearly-unbiased estimate of the attention matrix, uniform convergence, and decreasing variance of the approximation. We tested both models on our CNN-based models. According to our findings, neither model's optimizations reduced complexity since they focused on optimising the self-attention layer rather than the CNN layer, where the majority of the training time was spent, as evident by the Pytorch Profiler mapping of the resource usage, shown in \autoref{tbl:optimization}. To optimize the Learning Rate, we used the Pytorch built-in 1-Cycle Learning Rate optimizer. The 1-cycle schedule \citep{smith2017cyclical} operates in two phases, a cycle and a decay phase, with one iteration over the training data. 
In the cycle phase, the learning rate oscillates between a minimum value and a maximum value over some training steps. In the decay phase, the learning rate decays starting from the minimum value of the cycle phase. Using 1-cycle led to faster convergence of the model and better performance.

\begin{table}[htbp]
\caption{Resource Usage Using Pytorch Profiler \citep{paszke2019pytorch}}
\label{tbl:optimization}
\begin{center}
\small
\begin{tabular}{@{}l l   c  c  c     @{}}
\toprule
 & Operation Name & CUDA (ms) & CUDA (\%) & CUDA Total (ms)  \\\midrule
\multirow{9}{*}{} & Training step-CNN
 &  1.858  & 0.73 &  36.534     \\
 & Training step-Transformer &  15.831 & 3.14 & 138.421  \\
 & Attention::Linear &   7.958 & 6.24   &   110.619 \\
 & Attention::matmul & 5.775 & 2.28 & 23.401      \\
 & Attention::mm & 16.932  & 6.67 & 16.932      \\
 & \textbf{Cudnn-Convolution} & \textbf{73.958}  & \textbf{29.15} &  \textbf{73.958}   \\
 & Attention::Fused Dropout & 13.121  &  5.17 & 13.121    \\\bottomrule
\end{tabular}
\normalsize
\end{center}
\end{table}

\subsection{Segmentation: YOLO vs Detectron-2}
We experimented with YOLO-V5 \citep{glenn_jocher_2021_4679653}, a grid-based object recognition algorithm that divides pictures into grids, for segmentation. Each grid cell is responsible for detecting objects inside its own limits. YOLO is one of the most well-known object detection algorithms due to its speed and precision. YOLO is composed of three modules: a Backbone, a convolutional neural network that aggregates and forms image features at various granularities, a Neck, a series of layers that mix and combine image features before forwarding them to prediction, and a Head, which consumes Neck features and performs the box and class prediction steps. However, in our testing, YOLO-V5 frequently fails to capture marginal texts and other difficult content included in our dataset, although Detectron-2 excelled in most segmentation tasks.

\subsection{Decoding: beam search, greedy search, diverse beam search}

We evaluated three prominent decoding methods for post-correction. In most circumstances, the simplest technique is the best decoding: decoding your model output, which concatenates the most likely characters every time-step, and identifying the character with the greatest score each time, making Greedy Search optimal \citep{chickering2002optimal}. However, it may fail in other cases since it does not take into account all of the facts. A greedy algorithm's decision may be influenced by previous decisions, but it is unaware of potential future decisions. Another option is Beam Search (BS) \citep{wiseman2016sequence}, which creates and evaluates text candidate beams repeatedly. To begin, add an empty beamline and a matching score to the beam list. The method then iterates through all of the time-steps in the output. At each time step, we only save the best scoring beams from the previous time step. The beam width determines the number of beams to preserve (BW). Calculate the score for each of these beams at the current time step. In addition, each beam extends all possible characters from the alphabet and awards them a score. Return the best beam as a consequence of the latest time step. Beam search, on the other hand, generates lists of almost identical sequences, which is computationally inefficient and frequently fails to capture the intrinsic ambiguity of complicated AI tasks. To address this issue, we attempted Varied Beam Search (DBS) \citep{vijayakumar2016diverse}, a BS alternative that decodes a list of diverse outputs by optimising for a diversity-augmented objective. Our investigations revealed that Word Beam Search was the best fit for our situation.

\section{Results}\label{sec5}
This section presents the results of the conducted experiments for training the proposed OCR model on the constructed dataset. In this work, we trained on 100 000 images from the constructed dataset. Subsequently, the OCR model was trained using all 12 fonts available, with diacritics, and with both long and short sequences, achieving a CER of 4.46.

\section{Conclusion and Future Works}
In this research, we proposed a novel approach for transcribing historical Arabic manuscripts using an end to end Transformer architecture. We demonstrated various experiments conducted on different state-of-the-art models. In our future work, we aim to increase the model’s accuracy by training the model with more images from the constructed dataset since we believe that increasing the number of photos used to train the model will improve OCR accuracy considerably. Also, we aim to train on the larger variant of BEiT which we believe will greatly improve the model's predictions.

\section*{Acknowledgments}
The authors gratefully acknowledge the support of BibAlex for dedicating 64 dedicated node with two K80 Tesla GPU each.

\section*{Conflict of interests} The authors have declared that no conflict of interests interests exist.

\section*{Funding}
This research received no specific grant from any funding agency in the public, commercial, or not-for-profit sectors.

\section*{Materials Availability}
The dataset and code used to conduct the experiments in this paper will be made publicly available.

\bibliography{main}

\end{document}